\documentclass[journal]{IEEEtran}
\IEEEoverridecommandlockouts
\usepackage{cite}
\usepackage{booktabs}
\usepackage{amsmath,amssymb,amsfonts}
\usepackage{algorithmic}
\usepackage{graphicx}
\usepackage{textcomp}
\usepackage{amsmath,amssymb,amsfonts}
\usepackage{algorithmic}
\usepackage{array}
\usepackage{tabu}
\usepackage{tabularx}
\usepackage{svg}
\newcommand{\eatme}[1]{ }
\usepackage[inline]{enumitem}
\usepackage{float}
\usepackage{cite}
\usepackage{amsmath,amssymb,amsfonts}
\usepackage{algorithmic}
\usepackage{array}
\usepackage{tabu}
\usepackage{tabularx}
\usepackage{graphicx}
\usepackage{textcomp}
\usepackage{subcaption,siunitx,booktabs}
\usepackage{multirow}
\usepackage{ragged2e}
\usepackage{tabularx,tabulary}
\usepackage{adjustbox}
\usepackage{caption}
\renewcommand{\thetable}{\arabic{table}}   
\renewcommand{\thefigure}{\arabic{figure}}
\usepackage{xcolor}
\usepackage{geometry}
\usepackage{makecell}
\usepackage{fixltx2e}

\def\BibTeX{{\rm B\kern-.05em{\sc i\kern-.025em b}\kern-.08em
    T\kern-.1667em\lower.7ex\hbox{E}\kern-.125emX}}

\begin{document}
\pagenumbering{gobble}
% pretitle{\vspace*{-0.2in} \begin{center}\Huge\bfseries}
% DONT DO THIS^^^^

\title{\vspace{0.5cm} MTDT: A Multi-Task Deep Learning Digital Twin}
% \thanks{The work was supported in part by NSF CNS 1922782 and by the Florida Department of transportation (FDOT). The opinions, findings and conclusions expressed in this publication are those of the authors and not necessarily those of NSF or FDOT.}
%}

\newcommand{\linebreakand}{%
  \end{@IEEEauthorhalign}
  \hfill\mbox{}\par
  \mbox{}\hfill\begin{@IEEEauthorhalign}
}

\author{Nooshin Yousefzadeh, Rahul Sengupta, Yashaswi Karnati,  Anand Rangarajan, and Sanjay Ranka\\
%\textit{Department of Computer and Information Science and Engineering} \\
\textit{University of Florida, Gainesville, FL, USA} \\
}

\maketitle
%\footnote{\{nooshinyousefzad, rahulseng, yashaswikarnati,anandr,sranka\}@ufl.edu}
\begin{abstract}
Traffic congestion has significant impacts on both the economy and the environment. Measures of Effectiveness (MOEs) have long been the standard for evaluating traffic intersections' level of service and operational efficiency. However, the scarcity of traditional high-resolution loop detector data (ATSPM) presents challenges in accurately measuring MOEs or capturing the intricate spatiotemporal characteristics inherent in urban intersection traffic. To address this challenge, we present a comprehensive intersection traffic flow simulation that utilizes a multi-task learning paradigm. This approach combines graph convolutions for primary estimating lane-wise exit and inflow with time series convolutions for secondary assessing multi-directional queue lengths and travel time distribution through any arbitrary urban traffic intersection. Compared to existing deep learning methodologies, the proposed Multi-Task Deep Learning Digital Twin (MTDT) distinguishes itself through its adaptability to local temporal and spatial features, such as signal timing plans, intersection topology, driving behaviors, and turning movement counts. We also show the benefit of multi-task learning in the effectiveness of individual traffic simulation tasks. Furthermore, our approach facilitates sequential computation and provides complete parallelization through GPU implementation. This not only streamlines the computational process but also enhances scalability and performance.

\end{abstract}


\section{Introduction}
Traffic congestion in urban areas impacts efficiency, economics, environment, safety, and overall life quality. Measures of Effectiveness (MOEs) provide a benchmark of longitudinal perspectives to quantify the level of service and evaluate the performance of transportation systems. The employment of MOEs is also useful for optimizing the signal timing plan of intersections and the overall traffic flow accordingly. Common MOEs for urban traffic control include factors such as queue length, travel time, waiting time, number of stops, vehicle throughput, average speed, fuel efficiency and emissions, safety, equity, adaptability to demand, public satisfaction, accessibility, etc. Depending on the specific municipal goals and priorities, a combination of MOEs can be used to effectively evaluate the urban traffic control system. This study includes but is not limited to two commonly used MOEs: 1- Queue Length: Distance from the stop bar to the tail of the last vehicle stopped in any traffic lane, and 2- Travel Time: The time necessary to traverse any route connects any two points of interest.

% Microscopic simulators e.g., VISSIM, SUMO, and AIMSUN can use near to real trajectories, although they cannot be used for signal timing optimization purposes. It is Because they are normally time-consuming and cannot generalize to arbitrary topology and infrastructure of traffic urban roadways without painstakingly (re)drawing base maps and (re)setting traffic simulation parameters.

% compute the distribution of MOEs more accurately than macroscopic simulators using vehicles' effective locations at each time step so-called trajectories. Examples of such simulators include VISSIM, SUMO, and AIMSUN. However, they are normally time-consuming and cannot generalize to arbitrary topology and infrastructure of traffic urban roadways without painstakingly (re)drawing base maps and (re)setting traffic simulation parameters. Additionally, to find the optimal signal timing plan for intersections based on MOEs we need to run an exponentially increasing number of simulation scenarios that generate counterfactual traffic conditions while varying and combining the setting of different parameters. An example of such method is ReTime which parallelizes microscopic simulation instances on multiple processors.

Microscopic simulators e.g., VISSIM, SUMO, and AIMSUN fail to generalize to arbitrary topology and infrastructure of traffic urban roadways without painstakingly (re)drawing base maps and (re)setting traffic simulation parameters. Some data sources e.g., GPS, Video, BlueTooth, DSRC, etc. are normally sparse with privacy and ownership restrictions \cite{wolf2014applying}. Advanced sensing technologies such as induction loop detectors (ATSPM), make high-resolution loop detector traffic data widely available, although they fail to provide enough information for MOEs estimation by themselves. The computation of MOEs is specific to several tempo-spacial factors such as intersection characteristics(e.g., topology, lane configuration), traffic pattern (e.g., inflow traffic and turning-movement counts, signal timings, and traffic conditions (e.g., driving behaviors).

Existing analytical methods of estimating MOEs from ATSPM either use vehicle conservation input-output algorithms based on counting vehicle flow over advance and stop-bar detectors \cite{vigos2008real} or Lighthill’s shockwave-based methods that identify breakpoints in the detector actuation waveforms \cite{yao2019cycle}. Although, both of them are sensitive to the placement of detectors and limited to the localized training datasets. Machine learning methods use either GPS trace data \cite{fang2021ftpg} that are too coarse in recording dissipation trends, or ATSPM for cycle-wise MOEs which is not based on real-world traffic conditions due to the random vehicle flow and road networks without explicit left-turn buffers. Recently, some deep learning methods have employed geometric deep learning \cite{wright2019neural} and graph convolution networks \cite{karnati2021intertwin} to extract cycle-wise MOEs, although these approaches still fail to generalize the local traffic conditions and road geometries, also their reconstructed MOEs have as low resolution as approach level.

We propose a Multi-Task Deep Learning-based (MTDT) digital twin of intersections to address the aforementioned shortcomings. This model is trained on real-world data obtained from high-resolution induction loop detectors (ATSPM), capturing varying traffic patterns encompassing a wide range of configurable parameters, thus generating potentially viable counterfactual traffic scenarios. MTDT builds upon the previous work by Yousefzadeh et al. \cite{yousefzadeh2024graph}, extending it to create an accurate and reliable multi-purpose digital twin. The main difference between them is that \cite{yousefzadeh2024graph} introduces two individual deep learning-based digital twins that utilize graph neural networks to make a lane-wise estimation of either exit or inflow waveforms at any direction of an arbitrary intersection. We use a multi-task learning paradigm to combine graph convolution for learning primary lane-wise traffic waveforms with time series convolution for learning secondary multi-directional MOEs based on primary results for an arbitrary urban traffic intersection, all at the same time. While maintaining a straightforward design, our model emphasizes the advantages of multi-task learning in traffic modeling. Although additional complexities could potentially enhance feature learning, we have prioritized simplicity to underscore the benefits of multi-task learning in modeling traffic flow dynamics. By consolidating the learning process across multiple tasks, MTDT demonstrates reduced overfitting, increased efficiency, and enhanced effectiveness through the sharing of representations learned by different tasks. An overview of our model is depicted in Figure \ref{fig:intersection}, where orange-colored objects represent the inputs of the model, including stop-bar loop detector waveforms and signal timing plans (additionally are present driving behavior parameters and turning-movement counts features), while blue-colored objects represent the outputs of the model, including exit and inflow waveforms at each traffic lane, maximum queue length at each traffic phase, and travel time distribution at each traffic phase. Further details regarding the definition and computation of these MOEs are provided in Section \ref{datagen}. Traffic Data Generation.

% MTDT is trained on realistic traffic flow scenarios simulated by a microscopic simulator, with effective parameters perturbed within acceptable ranges. Parallel computation is utilized to generate a sizable dataset comprising approximately 3.6 million cycles, encompassing nine different intersections at a 5-second resolution. The real-world cycle lengths vary between 20 to 30 seconds. The model undergoes training on a diverse array of factual and counterfactual traffic scenarios, aimed at generalizing the concept of traffic flow dynamics across up to four-way urban signalized intersections. These intersections include exclusive left-turn buffers and actuated signal timing plans, catering to the need for adaptability in various municipal areas and at different times of the day. 

The following are the key contributions of our work:
\begin{itemize}
\item  MTDT jointly estimates fine-grained and lane-wise exit and inflow waveform time series, along with accurate travel time distribution and maximum queue lengths associated with every phase of the movement. The model's adaptability and responsiveness to the spatiotemporal intricacies of traffic dynamics are evident in its utilization of input parameters such as relevant driving behavior metrics, signal timing plans, turning movement counts, and signal timing plan parameters, among others.

% \item MTDT employs Attentional Graph Neural Networks (GATs) to tackle two primary tasks of modeling exit and inflow waveform time series. The output from primary tasks is then utilized to generate a multivariate one-dimensional time series, serving as inputs to Convolutional Neural Networks (CNNs) to perform two secondary tasks of modeling queue length and travel time measures.

\item Using our digital twins, it becomes feasible to accurately and reliably predict the impact of loop detector waveforms, signal timing plans, driving behavior parameters, and turning movement counts on lane-wise platoons and related MOEs. Importantly, this predictive capability is computationally faster than microscopic simulators, thanks to the utilization of GPU-based processors. The methodologies proposed in our digital twins incur $O(1)$ sequential computation and can be entirely parallelized using cost-effective GPU computation.

\item The findings of this study hold significance to extend to corridor and network level digital twins, and make benefits towards data-driven traffic control design and safety management, particularly in the context of smart intersections operating within highly dynamic environments.

\end{itemize}

We designed several experiments, whose findings aim to showcase the feasibility and efficacy of our multi-tasking approach in designing a deep learning-based and multifaceted digital twin for intersections. We introduce two variants of our digital twin and compare our model with them and with our previously introduced TQAM \cite{sengupta2021tqam} digital twin. The results of the experiments, underscore the versatility of our proposed digital twin, demonstrating its ability to not only generalize estimations across various intersection typologies and characteristics, but also successfully simulate ATSPM waveforms at lane level, and accurately estimate several Measures of Effectiveness (MOEs) in both time series and distribution forms at phase (lane-group) level for every approach (movement direction) within an urban intersection.

\begin{figure}[htbp]
        \centering
        \captionsetup{justification=raggedright,singlelinecheck=false}
        \includegraphics[scale=0.4]{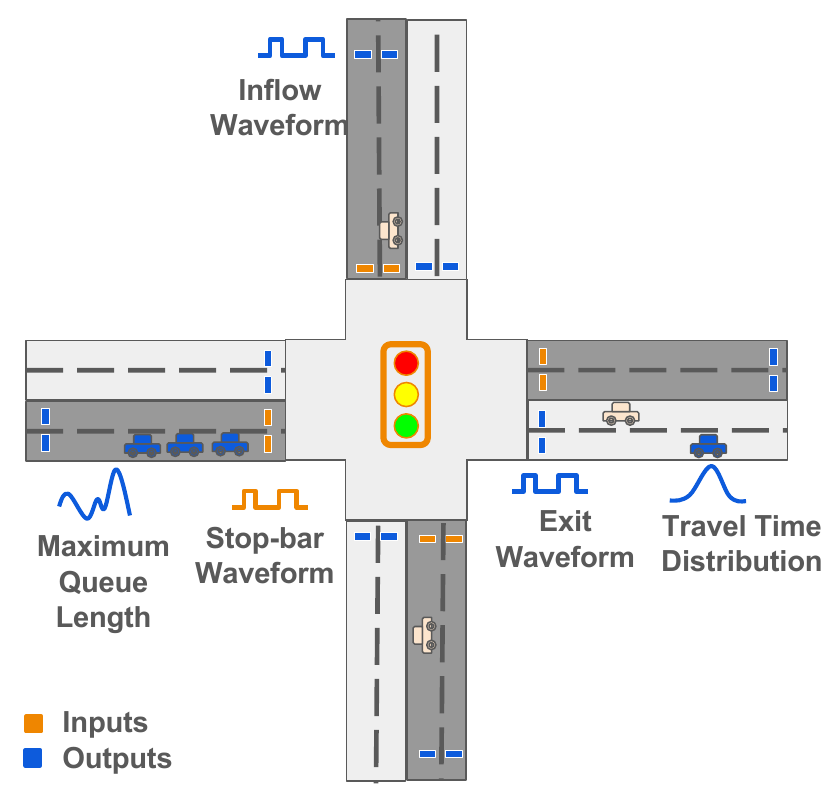}
        \renewcommand{\figurename}{Figure}
        \captionsetup{size=footnotesize}
\footnotesize
        \caption{\textbf{The inputs and outputs of Deep Multi-Task Learning Digital Twin} We simulate the ATSPM time series waveform along with waveform or histogram of several MOEs within 8-phase standard NEMA phasing intersections. MTDT simultaneously estimates downstream exit waveforms of every outflow lane in all directions and upstream inflow waveforms of every inflow lane in all directions with an estimation of instant maximum queue length and travel time distribution associated with each phase of movement for an intersection with arbitrary topology and traffic conditions.}
        \label{fig:intersection}
    \end{figure}

% We face some challenges in this study. The ATSPM data to the commonly used loop detector, stop bars. We used grid search to tune regularization hyperparameters in the customized loss function. For feasibility and safety reasons, the order of phases in the signal timing plan remains identical to the real field in the data generation. To preserve the spatial dependencies between lanes and still make the learning invariant to the specific topology of intersections, common lane-connectivity adjacencies are used in graph neural networks. 

% In generally, it is infeasible to vary the signal timing plans in the real-world, without proper consultation with the relevant stakeholders, such as the local public authorities and road users. Thus, we calculate MOEs in well-calibrated simulations. MOEs as reported in simulation are usually road segment-wise. We combine the MOEs of relevant road segments including the road segments immediately connected to the intersection, as well as those 1-hop away, and group them lane-group-wise (based on 8-phase NEMA phasing).

% Our model can also be utilized to identify customized effectiveness measures, offering a trade-off between existing MOEs at an intersection while evaluating various signal timing plans. 

The rest of the paper is organized as follows: Section~\ref{sec:related} provides an overview of related work, discussing various techniques employed for traffic state estimation. Section~\ref{proposedmodels} outlines the architecture of our proposed digital twin. Finally, in Section~\ref{expresults}, we assess and compare the performance of our model through a series of experiments.

\section{Problem Definition}
\label{problemdef}

This section extends the basic concepts and formal definitions introduced in our previous work to formulate our research problem in this study. For simplicity, the notations and terminologies are considered to be consistent. Detailed descriptions of the notations used in this study are presented in Table \ref{table:notation}. 

\noindent\textbf{Definition 1. Simulation Record:} This term refers to a record of a simulation run represented as a tuple of intersection ID, signal waveform $sig$, turn-movement counts vector $tmc$, driver behavior vector $drv$, lane-wise stop-bar loop detector waveforms $stp$, exit loop detector waveforms $ext$, and inflow loop detector waveforms $inf$, and multi-directional queue length waveforms $ql$, and travel time histograms $tt$.

\noindent\textbf{Definition 2. Simulation Graph:} This term refers to a single graph data item constructed using a standardized and generic structure, to represent a single traffic scenario occurring at a certain intersection. In this study, we construct two types of \textit{Simulation Graph} for each intersection $j$ from each single \textit{Simulation Record} $s$ each with its own attributes and structure to model exit traffic waveforms ($g_{s,j}=(V, E, X, Z)$) and inflow traffic waveforms ($g'_{s,j}=(V', E', X', Z)$). Both of them are directed bipartite graphs that connect each stop-bar loop detector to corresponding exit detectors in the node set $V$ or to inflow detectors in the node set $V'$. with edge features $Z = [sig_{j,w}, tmc_{j}, drv_{j}]$ at each graph defined as the concatenation of the signal timing plan time series, turning-movement counts vectors, and driving behavior parameters.

\begin{table}[ht]
% \scalebox{0.85}{
    \begin{adjustbox}{width=\columnwidth,center}
        \begin{tabulary}{1.0\textwidth}{|l|l|l|l|l|l|}
            \hline
            \textbf{Notation}&\textbf{Description}  &\textbf{Aggr}  &\textbf{Size}  &\textbf{Type}  \\ 
            \hline
            $j$&Intersection&-&1x9&String\\\hline
            $stp$&Waveform at Stop-bar detector  &5 sec  &48x80  &Integer, 0-8 \\\hline
            $ext$&Waveform at the exit of the intersection  &5 sec  &16x80  &Integer, 0-8 \\\hline
            $inf$&Waveform upstream the intersection  &5 sec  &12x80  &Integer, 0-8 \\\hline
            $ql$& Queue length time series&5 sec&8x80&Integer, 0-1200\\\hline
            $tt$& Travel time histogram&5 sec&8x200&Integer, 0-700\\\hline
            $sig$&Signal timing state information  &5 sec  &8x80  &Binary \\\hline
            $drv$& Driving behavior parameters  &2400 sec  &1x9 &Float, 0-30\\\hline
            $tmc$& Turning-movement counts ratio &2400 sec  &35x35  &Float, 0-1 \\\hline
            $w$& Size of the prediction window&-  &1  &Integer, 0-150 \\\hline
            $s$& Single simulation recorded by SUMO &5 sec&-&Multiple \\\hline
            $A$& Common lane-connectivity for $M_{ext}$ module&-&2x22&Binary\\\hline
            $A'$& Common lane-connectivity for $M_{inf}$ module&-&2x72&Binary\\\hline
            $v$& multivariate time series for queue length estimation &5 sec&1x7x80&-\\\hline
            $v'$& multivariate time series for travel time estimation &5 sec&1x7x80&-\\\hline
            $M_{ext}$& GAT module for exit waveforms estimation &-&-&torch.nn.Module\\\hline
            $M_{inf}$& GAT module for Inflow waveforms estimation&-&-&torch.nn.Module\\\hline
            $M_{ql}$& CNN module for queue length estimation&-&-&torch.nn.Module\\\hline
            $M_{tt}$& CNN module for travel time estimation&-&-&torch.nn.Module\\\hline
                        
        \end{tabulary}
    \end{adjustbox}
\caption{Summary of the notations and their definitions.}
\label{table:notation} 
\end{table}

\section{Proposed Models}
\label{proposedmodels}
The Multi-task Deep Learning Digital Twin (MTDT) is designed to grasp various facets of intersection-level traffic flow dynamics simultaneously. MTDT is jointly optimized to learn four sets of joint tasks. The general architecture of our digital twin is composed of two primary modules that yield base learning representations, and two secondary modules that use the representations learned by primary modules to learn their own tasks and improve the learning process of primary tasks as well by the sharing of their representations. For a comprehensive understanding of MTDT's architecture, refer to the overview depicted in Figure \ref{fig:overview}. Here, as described in Table \ref{table:notation}, $M_{ext}$ and $M_{inf}$ represent GAT modules that execute exit waveforms and inflow waveforms estimation respectively; while $M_{ql}$ and $M_{tt}$ represents CNN modules that execute queue length time series and travel time distribution estimation respectively. The output of primary tasks of size $16\times80$ for the exit task and $12 \times 80$ for the inflow task are aggregated to 8 lane groups (phase of movement) to the size $8 \times 80$ before being fed as input to the secondary tasks.

\begin{figure}[htbp]
        \centering
        \captionsetup{justification=RaggedRight, size=footnotesize}
        \includegraphics[scale=0.35]{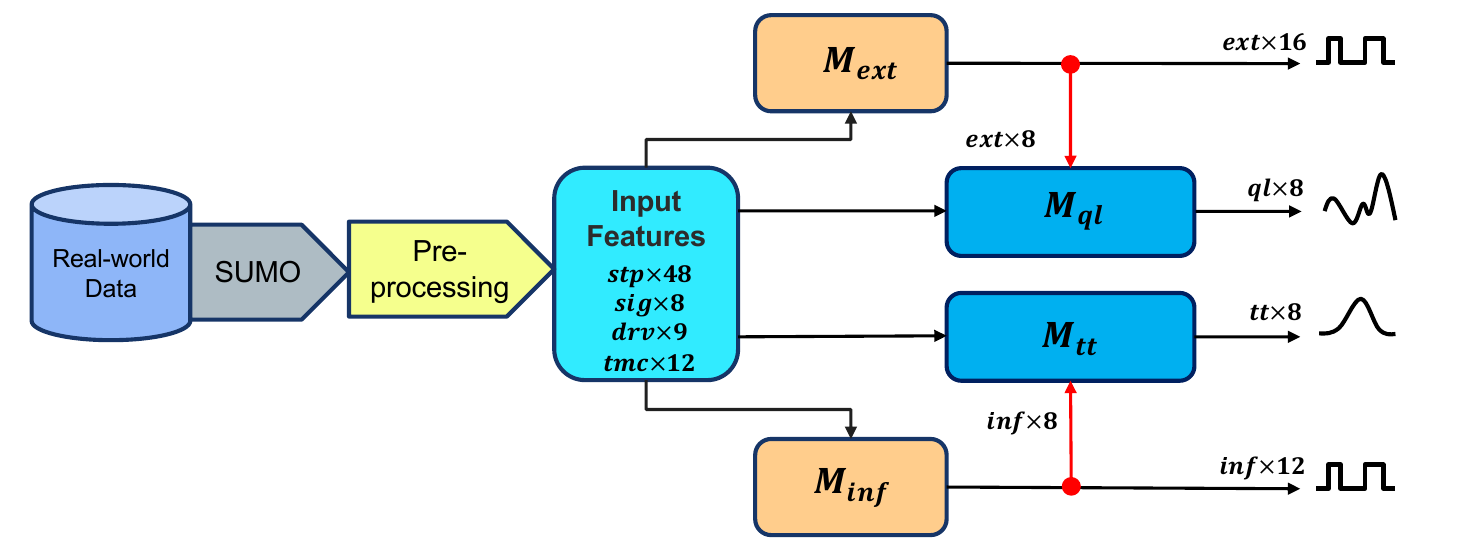}
        \footnotesize
        \caption{\textbf{Overview of the architecture of MTDT.} The architecture of the proposed Multi-componential digital twin framework contains two types of modules. GAT modules (light orange colored) execute primary tasks, which aggregated outputs are used as input to secondary tasks executed by CNN modules (dark blue colored). Red-colored arrows are activated only in inference mode.}
        \label{fig:overview}
    \end{figure}

The MTDT pipeline comprises two distinct types of modules: Graph Attention Networks (GATs) \cite{ velivckovic2017graph} modules, including $M_{ext}$ and $M_{inf}$, dedicated to the primary task of estimating loop detector waveforms, and Convolutional Neural Networks (CNNs) \cite{lecun1998gradient} modules, including $M_{ql}$ and $M_{tt}$, geared towards secondary tasks such as estimating Measures of Effectiveness (MOEs) like maximum queue length and travel time distribution. More specifically, GAT modules, namely $M_{ext}$ and $M_{inf}$, excel at precise, lane-specific estimations of waveform time series for vehicle platoons approaching and departing the target intersection. They meticulously consider instantaneous spatiotemporal factors like turning movement counts, driving behavior parameters, and signal timing plans. Subsequently, CNN modules, such as $M_{ql}$ and $M_{tt}$, utilize the outputs from primary modules alongside instantaneous spatiotemporal characteristics of the target intersection to estimate MOEs like maximum queue length and travel time distribution. This design ensures adaptability without restrictive assumptions, making the digital twin easily instantiated for various secondary tasks related to traffic dynamics.

The underlying principle of this architecture lies in extracting and leveraging valuable additional information from diverse aspects of the same phenomenon, resulting in enhanced performance and more comprehensive outcomes. Proper normalization techniques tailored to each task are employed to enhance the performance of both primary and secondary tasks collectively.

% Detailed information regarding the architecture and learning process within each module is elaborated upon in the subsequent subsections:

\subsection{Exit Module}
Exit subnetwork module $M_{ext}$ is an attentional graph neural network (GAT) with an imputation mechanism, as introduced in \cite{yousefzadeh2024graph} to perform the primary task of exit waveform time series $ext$ estimation. Here, the input data is fed in the form of \textit{Exit Simulation Graphs} with a common connectivity matrix and masked node features to connect a total number of 16 exit loop detectors' $ext$ waveform time series (for 4 outflow lanes for each approach) to the total number of 48 stop-bar loop detectors' $stp$ waveform time series as in a general topology of an arbitrary up to 4-way intersection. This module consists of a GATConv \cite{velivckovic2017graph} message passing layer following a single fully connected linear layer parameterized by weight matrix $B \in \mathbb{R}^{w \times N}$ and activated by a ReLU function to generate and propagate messages of hidden representation inflow lanes to any other connected outflow lanes within the intersection network. The attention mechanism in GATConv is a SINGLE-layer feed-forward neural network, parameterized by a weight vector $a \in \mathbb{R}^{N \times w}$ with ReLU non-linearity.
\[a_{ij}=ReLU(a^T[Wh_i\| Wh_j])\]
Through this technique, GATConv implicitly specifies and integrates the attention coefficients signifying the importance of each $stp$ waveform time series to each $ext$ waveform.
\[
\alpha_{ij}=Softamx_j(\frac{exp(e_{ij})}{\sum_{k\in N_i}exp(e_{ik}))}
\]

Where $N_i$ is the connected (neighboring node) lanes of each lane $i$ in the intersection's graph representation. A fully-connected layer activated by a ReLU function is applied on the output of the GATConv layer to impute missing (unobserved) time series waveforms for exit loop detectors $ext$ within reconstructed input feature matrix $\hat{X} \in \mathbb{R}^{N \times w}$. 

\subsection{Inflow Module}
Inflow subnetwork module $M_{inf}$ is an attentional graph neural network (GAT) with an imputation mechanism, as a simplified version of multi-layer graph neural network \cite{yousefzadeh2023comprehensive} solution introduced in \cite{yousefzadeh2024graph}. This module is responsible for performing the primary task of inflow waveform time series $inf$ estimation. The difference between $M_{ext}$ and $M_{inf}$ is that the same GATs module architecture is adopted to perform a different primary task. Here, GATConv implicitly specifies and integrates the attention coefficients denoting the importance $stp$ waveform time series to each $inf$ waveform using a different graph representation of an intersection that connects each of 12 inflow loop detectors' $ext$ waveform time series (for 3 inflow lanes for each approach) to the total number of 48 stop-bar loop detectors' $stp$ waveform time series as in a general topology of an arbitrary up to 4-way intersection.
In this setting, this GATs module can impute missing (unobserved) time series waveforms for exit loop detectors $inf$ within reconstructed input feature matrix $\hat{X} \in \mathbb{R}^{N' \times w}$.

\subsection{Queue Length Module}
Queue Length subnetwork module $M_{ql}$ is a multi-output 1D convolutional neural network (CNN) to perform the secondary task of maximum queue length waveform time series $ql$ estimation. Here, the input data is fed in the form of \textit{Multivariate Time series} $v_{s,j}$ of size $(8, 7, 80)$ each time step for each of 8 traffic phase, pair 7 variables including exit waveform time series $ext$ estimated by $M_{ext}$ with $stp$, driving behavior parameters $drv$, and signal timing plans $sig$ to capture appropriate and instant spatiotemporal characteristics of urban traffic intersections. In this setting, for simplicity reasons, only the three most effective driving behaviors i.e., accel, lcCooperative, minGap in traffic dynamic of intersection, based on SHAP analysis of the Average impact of features on the magnitude of estimation in \cite{yousefzadeh2024graph}, are included.
The architecture of the CNNs module consists of 3 Conv1d layers each followed by a MaxPool1d layer and flattening layer before being fed into a fully connected layer activated by a ReLU non-linearity to simultaneously generate a set of 8 outputs as an estimation to maximum queue length waveform time series for every phase of target standard NEMA eight-phasing intersection. 

\subsection{Travel Time Module} 
Queue Length subnetwork module $M_{tt}$ is a multi-output 1D convolutional neural network (CNN)  to perform the secondary task of travel time histogram $tt$ estimation. Here, the input data is fed in the form of \textit{Multivariate Time series} $v'_{s,j}$ of size $(8, 7, 80)$ each time step for each of 8 traffic phases, pair 7 variables including inflow waveform time series $inf$ estimated by $M_{inf}$ with $stp$, driving behavior parameters $drv$, and signal timing plans $sig$ to capture appropriate and instant spatiotemporal characteristics of urban traffic intersections. The architecture of this module remains the same as the $M_{ql}$ module.

\subsection{Loss Construction}
\label{loss}
Multi-task Deep Learning Digital Twin
(MTDT) is trained on numerous realistic traffic scenarios simulated across eight distinct intersections. It takes stop-bar high-resolution loop detector data, driving behavior parameters, signal timing plan parameters, and turning movement counts as inputs to generate exit and inflow time series waveforms for each lane of movement, alongside maximum queue length time series and travel time distribution for each phase of movement.

During the multi-tasking learning process in MTDT, information is extracted from four sources shown as $M = \{M_{ext}, M_{inf}, M_{ql}, M_{tt}\}$ learning modules.  These modules cater to four different yet interconnected tasks denoted as $T = \{ext, inf, ql, tt\}$ each yielding an output $\{y_t\}_{t \in T}$ and a computed loss value $\{l_t\}_{t \in T}$. The model undergoes simultaneous training across all tasks, with losses applied to facilitate learning across the entire spectrum. The final loss $ L = \sum_{t \in T} l_t$ is computed as the summation of partial loss values obtained from optimizing all primary and secondary learning tasks.

In the training phase, the learning process for both primary and secondary tasks operates independently while being supervised by log information generated by the SUMO micro-simulator for each target variable. The input information is represented either as \textit{Simulation Graphs} for GATs modules handling primary tasks or as \textit{Multivariate Time Series} for CNNs modules addressing secondary tasks. While, in the inference phase, the secondary tasks are facilitated by information derived from primary tasks involved in constructing their inputs (i.e., shown by the activation of red-colored arrows in Figure \ref{fig:overview}). The output of secondary modules in inference mode is generated by performing a 1D convolutional operation on the resulting 1D \textit{Multivariate Time Series}. It's worth noting that this study employs a simplified design to illustrate the benefit of this generic approach, more complex architectures with improved feature learning might lead to better results.

In the multi-tasking paradigm, consistency across the scale of outputs scales is important. In this experiment, a traffic scenario for a certain intersection includes several variables: $ext$ and $inf$ are time series with a range of 0-8 in the unit of number of vehicles, while $ql$ is time series with a range of 0-1200 in the unit of meter, and $tt$ is a histogram with range 0-700 in the unit of seconds. To optimize the generalization of the optimization scheme within our multi-task learning paradigm, we employ appropriate experimental normalization techniques tailored to each task to ensure equal weighting (prioritization) of partial loss values computed by different tasks. We use the min-max normalization technique for $y_{ext}$ and $y_{inf}$ and $y_{tt}$ and log normalization for $y_{ql}$. For each normalization and scaling formula, a counterpart denormalization function is utilized to transform $\{\hat{y}_t\}_{t \in T}$ back to the original scale before computing error values for estimations.

\[
\hat{y}_t =  Denorm_t(M_t(Norm_t(X_t)))
\]

The choice of loss function for each task also is aligned with the data type and task objectives. We use the mean squared error loss function for time series and cross-entropy loss function for histogram optimization treating the number of bins as the number of labels for its ordinal categorical variable. 

The model is trained on the train portion ($75\%$ split of data), and a grid search algorithm is employed to fine-tune regularization hyperparameters on the $15\%$ split of the validation portion. The model is evaluated on to the test partition and the error values are reported for the best-performing model across 100 epochs of training.

% In this study, the secondary sub-network modules are designed with as few parameters as possible and low computational overhead. This design ensures easy integration into the primary block, enhancing the efficiency of multi-task learning within the resulting model. 

In this study, we deliberately maintain a simplistic architecture to showcase the advantages of this learning paradigm in our application, introducing subtle additional complexities could potentially yield improved results through enhanced feature learning capabilities.

% although the definitions are slightly different for the different target variables. Since the maximum queue length should be considered in the computation of error values of queue length estimation, variants of MAE and RMSE so-called Mean Absolute Percentage Error (MAPE), and Root Mean Squared Percentage Error (RMSPE) give a better science of real error measurement for maximum queue length
% \[
% MAPE =  100/n \sum_{i=1}^n \frac{\hat{y}_t^i - y_t^i}{max(y_t)}
% \]

% \[
% RMSPE =  100 \times \sqrt{1/n \sum_{i=1}^n (\frac{\hat{y}_t^i - y_t^i}{y_t^i})^2}
% \]

\section{Traffic Data Generation}
\label{datagen}
Our dataset is based on over 400,000 simulation hours of SUMO data. A 9-intersection urban corridor located in a large metropolitan region in the United States of America was chosen. We reiterate the important points here; more details on the dataset are provided in \cite{yousefzadeh2024graph}.
 
% The intersections consist of four approaches (except one, which has 3 approaches). Induction loop detectors have been installed at stop-bar (at the intersection) and advance (around 100 meters upstream from the intersection) locations. Further, "exit" detectors have been placed at the beginning of the outgoing lanes. These are generally not seen in the real-world but have been simulated, in order to study the exiting traffic flow.

% \subsection{Traffic Generation}
We use real-world ATSPM (loop detector) data with sparse WEJO (GPS) probe trajectory data sources to generate an approximate Origin-Destination probability matrix. This provides insight into the probability of a vehicle originating from a specific location reaching various valid destinations. SUMO tool od2trips is used to generate route files for the simulation. 

Building upon the configuration parameters and the range of variation detailed in Supplementary Table 1 of \cite{yousefzadeh2024graph}, we augment our data generation process with additional processing. This enhancement enables the collection of several Measures of Effectiveness (MOEs) corresponding to the state of traffic flow across various simulated counterfactual traffic scenarios.

% We include counter-actual traffic scenarios as well. For this, we generate random vehicle flows, which vary between 0 to two times the average observed real-world flows. By doing so, we generate data with feasible yet unobserved turning-movement counts at the various intersections. %These scenarios may lead to partial or network-wide congestion due to cross-blocking, insufficient green times, etc.

For signal timing plans, we use Ring-and-Barrier operation with a common cycle length. Parameters such as the minimum/maximum green times, yellow times, and red times for the intersections are based on signal timing sheets. However, we do vary the common cycle lengths and barrier times. A random common cycle length is chosen between 120 seconds and 240 seconds. Once this is selected, random signal cycle offsets are chosen for each intersection separately. A random barrier time is chosen. Driving behavior parameters (such as acceleration, deceleration, emergency braking, minimum gap between vehicles, headway, lane-changing parameters, etc.) are also varied. Finally, with some additional processing, we get the outputs of several Measures of Effectiveness(MOEs) to measure the state of traffic flow generated at counterfactual scenarios.

% \subsection{Driving Behavior Variation}
% \subsection{Parallel Dataset Generation}
% The dataset was generated using HiperGator supercomputing resources using parallel computing libraries such as multiprocessing.

% While the dataset is largely processed based on \cite{yousefzadeh2024graph}, we do additional processing to get the outputs for the Measures of Effectiveness(MOEs) as described in the following:
\noindent\textbf{Queue Length.} The maximum queue length time series for each approach or traffic phase represents the instant magnitude of the maximum queue length in the unit of meter aggregated at a 5-second bucket resolution. In order to get the variation of queue length over time, we take the maximum queue length seen for a group of lanes associated with each phase of movement at a 5-second resolution. This is obtained directly by parsing SUMO log files and tracking queue lengths on a per-second basis. For the through-lanes (i.e. lanes serving the even number phases 2/4/6/8), we take the maximum queue lengths seen in 5 seconds across all (a) the 1-hop lanes serving the through-lanes and (b) 2-hop lanes w.r.t. the intersection. We then sum these two maximum queue lengths (a) and (b) to get the overall queue length. Broadly, this number gives the maximum possible queue length that any of the lanes belonging to the through-lanes, might experience.
For the left-lanes (i.e. lanes serving the odd number phases 1/3/5/7),  we take the maximum queue lengths seen in 5 seconds across all (a) the 1-hop lanes serving the left-lanes and (b) the left-most 2-hop lane w.r.t. the intersection.
We then sum these two maximum queue lengths (a) and (b) to get the overall queue length. The reason for choosing only the left-most 2-hop lane for (b) is that it is assumed that many vehicles that wish to take a left-turn, would have begun changing into the left-most lane in anticipation of the start of the left-turn buffer. Thus, the maximum queue length should take into account the left-most lane. 

\noindent\textbf{Intersection Travel Time.} The travel time histogram for each approach or traffic phase represents the frequency of vehicles that completed that traffic movement journey at the intersection groping into intervals of 5 seconds within a travel time range of 0 to 1000 seconds. This is obtained from SUMO logs, by processing its trajectory data at 1 second resolution. A vehicle is said to be within the proximity of the intersection if the vehicle is on a lane that is 1-hop or 2-hop away. The travel time here tracked is the amount of time the vehicle spends in the proximity of the intersection till it exits the intersection. For each vehicle, its travel time with respect to the intersections it crosses, is extracted. These travel times are collected for each intersection and each group of lanes associated with a phase of movement and a histogram is created. The histogram varies from 0 to 1000 seconds, in buckets of 5 seconds each. Each bucket counts how many vehicles had that travel time. Eg. a count of 7 vehicles in the bucket 45-50 s along phase 2 of intersection 5, means that 7 vehicles took between 45 and 50 seconds to cross the proximity of that intersection.

\section{Experimental Results}
\label{expresults}
In this section, we assess the performance of our proposed multi-tasking digital twin, referred to as MTDT. This MTDT model undergoes training on a dataset comprising 22,000 Simulation Records meticulously gathered from a diverse range of eight intersections. Additionally, we introduce two variants for comparison: MTDT-Single, trained on a dataset of 19,000 Simulation Records from a single intersection, and MTDT-MOE, a modified version of MTDT that excludes primary modules $M_{ext}$ and $M_{inf}$ from the pipeline, resulting in the output of only two sets of MOEs variables, namely $ql$ and $tt$, instead of the full set of variables. Lastly, we include TQAM \cite{sengupta2021tqam} an RNN Auto-encoder as a baseline deep learning-based digital twin previously introduced in our research for the estimation of maximum queue length.

% As already discussed, knowledge transfer induction generated by the multi-task learning paradigm become beneficial in MTDT and allows the this single shared deep-learning model to simultaneously reconstruct time series or distribution of $4 \times 8 \times 9$ MOE variables for all 8 phases, in addition to $11 \times 9$ Exit waveform time series and $12 \times 12$ Inflow waveform time series, both for a total number of 9 mixed intersections. 

We employ widely used metrics to assess our model's performance, including Mean Absolute Error (MAE), Root Mean Square Error (RMSE), and Normalized Root Mean Squared Error (NRMSE). For travel time distribution estimation, we dissect error measures for the 60th, 75th, 85th, and 90th percentiles separately. This breakdown allows us to scrutinize the model's performance across various traffic flow dynamics within each percentile. When estimating queue length waveform time series, we opt for NRMSE over RMSE to ensure error values are relative to the scale of data. Additionally, to mitigate scale-related biases, we segment our dataset based on the magnitude of the maximum queue length logged by each \textit{Simulation Records} and report error values for each group individually.

% Furthermore, another experiment from a different perspective evaluates our digital twin's performance under scenarios where significantly different green times occur at major movement directions (phase 2/6) of intersections

Table. \ref{table:accuracy} shows our proposed digital twin (MTDT) has the potential to effectively learn the joint estimation of traffic flow and MOE estimation at any intersection regardless of its lane design and configuration (topology). Although MTDT-Single is trained and evaluated at a single intersection, the error values are fairly comparable to that for MTDT except for maximum queue length estimation. This is because the formation of queues at the location of each intersection follows a more complex pattern involving several intersection-level factors. The benefit of multi-task learning can be proved by comparing the maximum queue length estimation by MTDT-MOE which is designed to estimate MOEs exclusively and MTDT-Single. Supporting the performance made for queue length by the output of primary tasks, results in better estimation results.

As shown in MTDT results, the minimal marginal increase in error values for all variables remains minimal relative to the aggregation level indicating consistent reliability and robustness in the model's performance across varying data resolutions. We also note the model exhibits enhanced performance in estimating travel time for lower quantiles of the data distribution while showing a slight improvement, particularly at the 90th quantile. This suggests that our designed digital twin is also particularly adept at capturing relatively rare or extreme patterns.

% As exit and inflow time series are aggregated into larger bucket sizes, the increase in error values remains minimal relative to the aggregation level. This indicates consistent reliability and robustness in the model's performance across varying data resolutions. For all models, lower Confidence Intervals (CI) denote greater precision and certainty in exit waveform estimation compared to inflow waveforms. Similarly, the certainty in travel time distribution estimation by MTDT-Single slightly surpasses that of MTDT. Additionally, we note that the model exhibits enhanced performance in estimating travel time for lower quantiles of the data distribution while showing a slight improvement, particularly at the 90th quantile. This suggests that our designed digital twin is also particularly adept at capturing relatively rare or extreme patterns.

\begin{table*}[htbp]
    \centering
    \scalebox{0.8}{
    \begin{adjustbox}{width=\textwidth,center}
    \begin{tabulary}{\textwidth}{|L |L |L |L |}

             \hline
                \multirow{2}{*}{\textbf{Aggregation}} & 
                \multirow{2}{*}{\textbf{MTDT-Single}}&
                \multirow{2}{*}{\textbf{MTDT}}&
                \multirow{2}{*}{\textbf{MTDT-MOE}} \vspace{2mm}\\

            \hline
            \multicolumn{4}{|c|}{\textbf{Maximum Queue Length Estimation}}\\
            \hline
            \hfil &\hfil \textbf{$MAE|RMSE$}&\hfil \textbf{$MAE|RMSE$}&\hfil \textbf{$MAE|RMSE$}\\
            \hline
            5-second buckets &  $4.8084|0.0885$ &  9.6114$|$0.04621&  $9.0321|0.0297$  \\
            \hline
            10-second buckets &   $9.3787|0.08410$ &$18.9512|0.08118$ &  $17.8554|0.0513$ \\
            \hline
            15-second buckets &   $13.8788|0.0973$ &$27.9748|0.1012$  &  $26.4127|0.0674$  \\
            \hline
            20-second buckets &   $18.2466|0.0907$ &$37.1869|0.1003$  &  $35.1794|0.0688$\\
            \hline
            \multicolumn{4}{|c|}{\textbf{Travel Time Distribution Estimation}}\\
            \hline
            \hfil &\hfil \textbf{$MAE|RMSE|95\%CI$}&\hfil \textbf{$MAE|RMSE|95\%CI$}&\hfil \textbf{$MAE|RMSE|95\%CI$}\\
            \hline
            60th percentile &   $5.50946|17.744|\mp 34.7782$ &$5.5332|17.7440|\mp 34.7782$  &  $4.8024|15.3052| \mp 29.9981$   \\
            \hline
            75th percentile &   $5.7687|17.7987|\mp 34.8854$& $5.7592|18.3542|\mp 35.9742 $ &  $4.8387|15.2256| \mp 29.8421$  \\
            \hline
            85th percentile &   $4.7090|13.0162|\mp 25.5117$  &$5.7298|18.1125|\mp 35.5005$  &  $4.6844|14.8253| \mp 29.0575$    \\
            \hline
            90th percentile &   $4.1331|11.7298|\mp 22.9904$& $5.0685|16.3674|\mp 32.0801$   &  $3.7703|12.3752| \mp 24.2553$    \\
            \hline
            \multicolumn{4}{|c|}{\textbf{Exit Waveform Estimation}}\\
            \hline
            \hfil &\hfil \textbf{$MAE|RMSE|95\%CI$}&\hfil \textbf{$MAE|RMSE|95\%CI$}&\hfil \textbf{$MAE|RMSE|95\%CI$}\\
            \hline
            5-second buckets &  $0.0979|0.2995|\mp 0.5870$  &$0.1020|0.3189|\mp 0.6250$  &  -  \\
            \hline
            10-second buckets &   $0.1758|0.5083|\mp 0.9962$  &$0.1844|0.5575|\mp 1.0927$  &  -   \\
            \hline
            15-second buckets &  $ 0.2477|0.6977|\mp 1.3674$  &$0.2607|0.7761|\mp 1.5211$  &  -   \\
            \hline
            20-second buckets &   $0.3126|0.8683|\mp 1.7018$  &$0.3326|0.9775|\mp 0.9775$  &  -   \\
            \hline
            \multicolumn{4}{|c|}{\textbf{Inflow Waveform Estimation}}\\
            \hline
            \hfil &\hfil \textbf{$MAE|RMSE|95\%CI$}&\hfil \textbf{$MAE|RMSE|95\%CI$}&\hfil \textbf{$MAE|RMSE|95\%CI$}\\
            \hline
            5-second buckets &  $0.1145 | 0.3456 | \mp 0.6773 $ &$ 0.1284 | 0.3785 | \mp .7418$  &  -  \\
            \hline
            10-second buckets & $0.2119 | 0.6005 | \mp 1.1769$ &$0.2385 | 0.6729 | \mp 1.3188$  &  -   \\
            \hline
            15-second buckets & $ 0.3018 | 0.8287 | \mp 1.6242$ &$0.3397 | 0.9387 | \mp 1.8398$  &  - \\
            \hline
            20-second buckets &  $0.3837 | 1.0340 | \mp 2.0266$  &$0.4367 | 1.1844 | \mp 2.3214$  &  - \\
            \hline
    \end{tabulary}
    \end{adjustbox}}
    \caption{\textbf{Performance of MTDT.} The performance of our proposed digital twin is assessed across various aggregation resolutions of maximum queue length and at different quantiles of the travel time distribution. MTDT is compared to its variant (MTDT-Single and MTDT-MOE). This comparison is conducted in terms of NRMSE, RMSE, and MAE, measured in vehicle-per-bucket for Exit and Inflow waveforms, meter-per-bucket for maximum queue length, and vehicle-per-quantile for travel time distribution. Moreover, the level of confidence for estimating the relevant variable across various resolution levels is shown as 95\% confidence intervals (CI).}
    \label{table:accuracy}
\end{table*}

For the multifaceted and fine-grained results, our designed multi-tasking simulation stands as unique to its counterparts, thus not directly comparable to existing baselines. However, we attempt to contrast our findings with those obtained by TQAM designed to make an accurate estimation of queue length estimation in a specific approach of a specific topology of target intersection. To achieve this, we partition our data into the same queue length buckets as reported for this model and compare error values in terms of (percentage) Mean Absolute Error (MAE) in Table \ref{table:QL}. The partitioning encompasses two groups for each of the low (L), medium (M), and high (H) ranges of maximum queue length. These groups, ranging from L1 to H2, encompass 28\%, 34\%, 19\%, 11\%, 5\%, and 3\% of data samples, respectively. Our observations indicate that both MTDT and MTDT-Single exhibit comparable performance to TQAM. Specifically, MTDT outperforms TQAM in certain partitions of data where very short (0-40 meters) or long (above 200 meters) queue length tails are recorded, while MTDT-Single consistently performs equivalently or better than TQAM across nearly all partitions. 

% Supplementary Table 1 presents the error values calculated for individual intersections, with their simulation records split based on Maximum Queue Length buckets. The consistency in performance indicates that our model effectively generalizes the learning task across heterogeneous topologies and characteristics of intersections.

\begin{table*}[!htb]
\scalebox{0.9}{
\begin{tabular}{|lllllll|}
\hline
\begin{tabular}[c]{@{}l@{}}Max. Queue \\Length  (meter):\end{tabular}& 
\begin{tabular}[c]{@{}l@{}}\textbf{L1}\\0-40 m\end{tabular}&
\begin{tabular}[c]{@{}l@{}}\textbf{L2}\\40-80 m\end{tabular}&
\begin{tabular}[c]{@{}l@{}}\textbf{M1}\\80-120 m\end{tabular}&
\begin{tabular}[c]{@{}l@{}}\textbf{M2}\\120-160 m\end{tabular}&
\begin{tabular}[c]{@{}l@{}}\textbf{H1}\\160-200 m\end{tabular}&
\begin{tabular}[c]{@{}l@{}}\textbf{H2}\\200+ m \end{tabular}\\
\hline

\hline
\begin{tabular}[c]{@{}l@{}}\textbf{MTDT} \end{tabular} &
3.02 (7.5\%)&
6.00  (7.5\%)&
8.65 (7.2\%) &
11.34 (7.0\%)&
16.46 (8.2 \%)&
15.26 (1.5\%)\\

\hline
\begin{tabular}[c]{@{}l@{}}\textbf{MTDT-Single} \end{tabular} &
2.94 (7.3 \%) &
5.42 (6.7 \%)&
7.27 (6.0 \%)  &
9.16 (5.7 \%)&
11.05 (5.5 \%)&
9.80 (0.9 \%)\\

\hline
\textbf{TQAM}&2.90 (7.3\%) & 5.50 (6.9\%) & 7.00 (5.8\%) & 9.00 (5.6\%) & 11.60 (5.8\%) &18.10 (1.8\%) \\ \hline
\end{tabular}}
\caption{\textbf{Evaluation at various maximum queue length ranges} A comparative analysis is conducted between MTDT, MTDT-Single, and TQAM for queue length waveform time series estimation across different queue length buckets. The assessment is based on MAE and the percentage of MAE relative to the upper range within each bucket, measured in meters per 5-second time resolution.}
\label{table:QL}
\end{table*}

We further devise an experiment to assess the performance of our proposed digital twin under varying percentages of green time allocation to the major direction of the intersection. In this experimental setup, we focus on Phase 2/6 in the corridor-through phase, serving the \textit{major} direction of traffic movement for all intersections. We define low, medium, and high percentages of this green time as 45-50\%, 60-75\%, and 75-90\% of the total duration of the simulation scenario. However, the individual cycle lengths may vary due to actuated ring-and-barrier operation responding to the cycle-wise incoming traffic.  
After categorizing simulation records into these three groups, we evaluated the performance of MTDT and MTDT-Single across all output variables, as presented in Table \ref{table:GT}. The findings indicate that the performance of our digital twin is fairly consistent. It excels in queue length estimation under scenarios with shorter green time allocated to the major direction, whereas travel time estimation performs better with a larger green time allocation. 

% Supplementary Table 2 presents the error values calculated for individual intersections, with their simulation records split based on Corridor-Through Green Time Percentage buckets. The consistency in performance indicates that our model effectively generalizes the learning task across heterogeneous topologies and characteristics of intersections.

% In the supplementary document with this paper, we include two additional tables. In them, we see that the results (split based on Maximum Queue Length buckets and on Corridor-Through Green Time Percentage buckets) for various intersections when calculated separately, show consistent performance, indicating that our model is able to generalize across heterogeneous intersections.

% % \scalebox{0.85}{
%     \begin{adjustbox}{width=\columnwidth,center}
%         \begin{tabulary}{1.0\textwidth}{|l|l|l|l|l|l|}

\begin{table}[!htb]
\scalebox{0.75}{
\begin{tabulary}{1.0\columnwidth}{llllll}
\hline
\begin{tabular}[c]{@{}l@{}}Green Time (percentage):\end{tabular}&
&
\begin{tabular}[c]{@{}l@{}}\textbf{Low}\\45-60 \%\end{tabular}&
\begin{tabular}[c]{@{}l@{}}\textbf{Medium}\\60-75 \%\end{tabular}&
\begin{tabular}[c]{@{}l@{}}\textbf{High}\\75-90 \%\end{tabular}\\

\textbf{Task} & \textbf{Error Measure}&&&\\

\hline
\begin{tabular}[c]{@{}l@{}@{}} \\ \textbf{MTDT-Single}\end{tabular}\\
\hline

\begin{tabular}[c]{@{}l@{}}Exit Waveform \end{tabular} &
\begin{tabular}[c]{@{}l@{}}MAE\\RMSE \end{tabular} &
\begin{tabular}[c]{@{}l@{}}0.0106\\ 0.3556\end{tabular} &
\begin{tabular}[c]{@{}l@{}}0.0963\\ 0.3121\end{tabular} &
\begin{tabular}[c]{@{}l@{}}0.0936\\ 0.3439 \end{tabular} \\
\hline
\begin{tabular}[c]{@{}l@{}}Inflow Waveform \end{tabular} &
\begin{tabular}[c]{@{}l@{}}MAE\\RMSE \end{tabular} &
\begin{tabular}[c]{@{}l@{}}0.1518\\ 0.1199\end{tabular} &
\begin{tabular}[c]{@{}l@{}}0.3632\\ 0.320\end{tabular} &
\begin{tabular}[c]{@{}l@{}}0.4113\\ 0.3430\end{tabular} \\
\hline
\begin{tabular}[c]{@{}l@{}}Maximum Queue Length \end{tabular} &
\begin{tabular}[c]{@{}l@{}}MAE\\NRMSE \end{tabular} &
\begin{tabular}[c]{@{}l@{}}6.0780\\ 0.2071\end{tabular} &
\begin{tabular}[c]{@{}l@{}}5.9700\\ 0.1392\end{tabular} &
\begin{tabular}[c]{@{}l@{}}4.5635\\ 0.0778\end{tabular} \\
\hline
\begin{tabular}[c]{@{}l@{}}Travel Time Distribution  \end{tabular} &
\begin{tabular}[c]{@{}l@{}}MAE\\RMSE \end{tabular} &
\begin{tabular}[c]{@{}l@{}}1.4173\\ 4.9942\end{tabular} &
\begin{tabular}[c]{@{}l@{}}2.8246\\ 9.7389\end{tabular} &
\begin{tabular}[c]{@{}l@{}}3.1870\\ 10.827\end{tabular} \\

\hline
\begin{tabular}[c]{@{}l@{}@{}} \\ \textbf{MTDT}\end{tabular}\\
\hline

\begin{tabular}[c]{@{}l@{}}Exit Waveform \end{tabular} &
\begin{tabular}[c]{@{}l@{}}MAE\\RMSE \end{tabular} &
\begin{tabular}[c]{@{}l@{}}0.0905\\0.3288\end{tabular} &
\begin{tabular}[c]{@{}l@{}}0.09108\\ 0.0800\end{tabular} &
\begin{tabular}[c]{@{}l@{}}0.0926\\ 0.3052\end{tabular} \\
\hline
\begin{tabular}[c]{@{}l@{}}Inflow Waveform \end{tabular} &
\begin{tabular}[c]{@{}l@{}}MAE\\RMSE \end{tabular} &
\begin{tabular}[c]{@{}l@{}}0.1204\\ 0.4052\end{tabular} &
\begin{tabular}[c]{@{}l@{}}0.1151\\ 0.3811\end{tabular} &
\begin{tabular}[c]{@{}l@{}}0.1101\\ 0.3438\end{tabular} \\
\hline
\begin{tabular}[c]{@{}l@{}}Maximum Queue Length \end{tabular} &
\begin{tabular}[c]{@{}l@{}}MAE\\NRMSE \end{tabular} &
\begin{tabular}[c]{@{}l@{}}12.105\\ 0.4243\end{tabular} &
\begin{tabular}[c]{@{}l@{}}11.946\\ 0.2780\end{tabular} &
\begin{tabular}[c]{@{}l@{}}8.6872\\ 0.1557\end{tabular} \\
\hline
\begin{tabular}[c]{@{}l@{}}Travel Time Distribution  \end{tabular} &
\begin{tabular}[c]{@{}l@{}}MAE\\RMSE \end{tabular} &
\begin{tabular}[c]{@{}l@{}}0.0915\\ 8.4421\end{tabular} &
\begin{tabular}[c]{@{}l@{}}2.3671\\ 9.0937\end{tabular} &
\begin{tabular}[c]{@{}l@{}}3.0768\\ 10.735\end{tabular} \\

\hline
\end{tabulary}
}
\captionsetup{width=.5\textwidth}
\caption{\textbf{Evaluation at various green time allocations.} The performance of the model is evaluated at different plit of data based on corridor-through green time percentage buckets.}

\label{table:GT}

\end{table}

\section{Related Work}
\label{sec:related}
Existing methods of traffic flow modeling in the past literature can be categorized into linear, non-linear, machine learning, and hybrid models. With advancements in high-resolution induction loop detectors in intelligent transportation systems (ITSs), a large number of deep neural network models have been proposed for real-time and short-term traffic flow prediction raised by industries, academia, and government.
In the realm of linear models, \cite{chen2018localized} proposed a localized (dynamic weight matrix) space-time autoregressive (LSTAR) model for real-time traffic prediction, \cite{emami2019using} developed a Kalman filter (KF) for predicting real-time traffic flow of connected vehicles at urban arterials, \cite{ma2017short} combines KF and adaptive weight allocation, and \cite{safarinejadian2015consensus} formulates a consensus filter-based distributed variational Bayesian (CFBDVB) algorithm for density approximation of traffic flow and average traffic speed in a freeway. 

In the realm of non-linear models, multi-kernel SVM \cite{ling2017short} and weighted k-nearest neighbor \cite{xia2016distributed} with MapReduce algorithm on Hadoop are introduced. When it comes to hybrid models, ensemble empirical mode decomposition with adaptive noise \cite{tian2019hybrid} and bidirectional RNN combined with road crossing vector coding \cite{zhao2019traffic} can be mentioned.  
 % \cite{ Multiple measures-based chaotic time series for traffic flow prediction based on Bayesian theory} combined radial basis function (RBF) neural network with Bayesian estimation theory to predict short-term traffic flow out of its chaotic characteristics.
Machine learning models in this line of research are deep learning frameworks based on the temporal convolutional network (TCN)   
\cite{zhao2019deep}, parallel computing of deep belief networks (DBNs) \cite{zhao2019parallel}, deep characteristic learning using DBN with the multi-objective particle swarm optimization (PSO) algorithm \cite{li2019day}. \cite{duan2018improved} and \cite{ma2018short} developed a deep hybrid neural network combined with CNN and LSTM using trajectory data for better accuracy and robustness of forecasting performance, while \cite{yousefzadeh2024graph} combined GCN and attention mechanism for lane-wise and topology invariant traffic flow estimation. 

Some literature integrates operation performance forecasting in traffic signal control of intersections and networks. For example, \cite{nematichari2022evaluating} makes use of graph mining and trajectory data mining methods to provide interactive evaluation and exploration of the various MOEs, while \cite{du2023safelight} proposed a safety-enhanced residual reinforcement learning method (SafeLight) to integrate safety conditions in traffic signal control by integrating knowledge using multi-objective loss function and reward shaping, and \cite{jamal2020intelligent} proposed a metaheuristic-based method for intelligent traffic control using Genetic Algorithm (GA) and differential evolution (DE) to improve the level of service (LOS) i.e. delay optimization of intersection through optimizing the signal timing plan.

\section{Conclusions and Future Work}
\label{conclusion}
In this paper, we demonstrate the effectiveness of multi-tasking as a learning paradigm for deep learning-based digital twins. We introduce a Multi-Task Deep Learning Digital Twin that utilizes multiple convolutions over graphs and time series. This approach can successfully perform a variety of related traffic simulation tasks for urban intersections with arbitrary topologies and characteristics. Our findings highlight the benefits of the multi-task learning approach in the design of our digital twin, not only for secondary tasks whose inputs rely on the output of primary tasks but also for primary tasks to generalize and enhance their performance through shared learned representations and joint optimization. Furthermore, we design two experiments to specifically illustrate how the model's performance varies across different partitions of data when simulation scenarios are segmented based on signal timing plans or maximum recorded queue length. The results underscore the consistency and robustness of our approach across different experiments. Moreover, we discuss the lack of a baseline due to the uniqueness and comprehensiveness of our approach. However, we show it still outperforms in individual tasks compared to some previously introduced digital twins specialized for specific tasks.

Looking ahead, there are several aspects of future work in this research that we briefly discuss in the following. 

% In this approach, we use lane-wide inflow time series waveforms $inf$ estimated by the graph-based digital twin $G_{inf}$ as vehicles approach the target intersection from various directions, we construct the Origin-Destination (O-D) matrix. Subsequently, we employ the od2trips tool of SUMO to generate route files and evaluate each traffic scenario using a variety of Measures of Effectiveness (MOEs). An Optimum signal timing plan can be viewed as the most favorable trade-off among desired MOEs, considering a range of traffic conditions such as signal timing plans and driving behaviors. We consider this topic a significant avenue for our future research. %that will be eventually found using a Monte Carlo tree search (MCTS) \cite{browne2012survey}.

MTDT can serve as a valuable tool to assist traffic decision-makers in comprehending the potential impact of alterations to signal timing plans and proposed changes to road geometry. Additionally, it can be seamlessly integrated into signal timing optimization frameworks or expanded from isolated intersections to corridors and networks which will be a focal point of our future endeavors.

% Moreover, this approach can be expanded from isolated intersections to corridors and networks, paving the way for traffic signal timing plan optimization, which will be a focal point of our future endeavors. Moreover, our model holds promise for integration into MapReduce frameworks, facilitating distributed training and parallel prediction of traffic flow.
%or adaptive traffic signal control softwares (e.g, TRANSYT-7F, SCOOT) based on Robertson platoon dispersion learning for signal timing optimization at the level of corridor or network.

\section{Acknowledgments}
The work was supported in part by NSF CNS 1922782. The opinions, findings, and conclusions expressed in this publication are those of the authors and not necessarily those of NSF. The authors also acknowledge the University of Florida Research Computing for providing computational resources and support that have contributed to the research results reported in this publication.

% \section*{References}

\bibliographystyle{abbrv}
\bibliography{ref}

\begin{thebibliography}{10}

\bibitem{chen2018localized}
J.~Chen, D.~Li, G.~Zhang, and X.~Zhang.
\newblock Localized space-time autoregressive parameters estimation for traffic flow prediction in urban road networks.
\newblock {\em Applied Sciences}, 8(2):277, 2018.

\bibitem{du2023safelight}
W.~Du, J.~Ye, J.~Gu, J.~Li, H.~Wei, and G.~Wang.
\newblock Safelight: A reinforcement learning method toward collision-free traffic signal control.
\newblock In {\em Proceedings of the AAAI Conference on Artificial Intelligence}, volume~37, pages 14801--14810, 2023.

\bibitem{duan2018improved}
Z.~Duan, Y.~Yang, K.~Zhang, Y.~Ni, and S.~Bajgain.
\newblock Improved deep hybrid networks for urban traffic flow prediction using trajectory data.
\newblock {\em Ieee Access}, 6:31820--31827, 2018.

\bibitem{emami2019using}
A.~Emami, M.~Sarvi, and S.~Asadi~Bagloee.
\newblock Using kalman filter algorithm for short-term traffic flow prediction in a connected vehicle environment.
\newblock {\em Journal of Modern Transportation}, 27:222--232, 2019.

\bibitem{fang2021ftpg}
M.~Fang, L.~Tang, X.~Yang, Y.~Chen, C.~Li, and Q.~Li.
\newblock Ftpg: A fine-grained traffic prediction method with graph attention network using big trace data.
\newblock {\em IEEE Transactions on Intelligent Transportation Systems}, 23(6):5163--5175, 2021.

\bibitem{jamal2020intelligent}
A.~Jamal, M.~Tauhidur~Rahman, H.~M. Al-Ahmadi, I.~Ullah, and M.~Zahid.
\newblock Intelligent intersection control for delay optimization: Using meta-heuristic search algorithms.
\newblock {\em Sustainability}, 12(5):1896, 2020.

\bibitem{karnati2021intertwin}
Y.~Karnati, R.~Sengupta, and S.~Ranka.
\newblock Intertwin: Deep learning approaches for computing measures of effectiveness for traffic intersections.
\newblock {\em Applied Sciences}, 11(24):11637, 2021.

\bibitem{lecun1998gradient}
Y.~LeCun, L.~Bottou, Y.~Bengio, and P.~Haffner.
\newblock Gradient-based learning applied to document recognition.
\newblock {\em Proceedings of the IEEE}, 86(11):2278--2324, 1998.

\bibitem{li2019day}
L.~Li, L.~Qin, X.~Qu, J.~Zhang, Y.~Wang, and B.~Ran.
\newblock Day-ahead traffic flow forecasting based on a deep belief network optimized by the multi-objective particle swarm algorithm.
\newblock {\em Knowledge-Based Systems}, 172:1--14, 2019.

\bibitem{ling2017short}
X.~Ling, X.~Feng, Z.~Chen, Y.~Xu, and H.~Zheng.
\newblock Short-term traffic flow prediction with optimized multi-kernel support vector machine.
\newblock In {\em 2017 IEEE Congress on Evolutionary Computation (CEC)}, pages 294--300. IEEE, 2017.

\bibitem{ma2018short}
D.~Ma, B.~Sheng, S.~Jin, X.~Ma, and P.~Gao.
\newblock Short-term traffic flow forecasting by selecting appropriate predictions based on pattern matching.
\newblock {\em IEEE Access}, 6:75629--75638, 2018.

\bibitem{ma2017short}
M.~Ma, S.~Liang, H.~Guo, and J.~Yang.
\newblock Short-term traffic flow prediction using a self-adaptive two-dimensional forecasting method.
\newblock {\em Advances in Mechanical Engineering}, 9(8):1687814017719002, 2017.

\bibitem{nematichari2022evaluating}
A.~Nematichari, T.~Pechlivanoglou, and M.~Papagelis.
\newblock Evaluating and forecasting the operational performance of road intersections.
\newblock In {\em Proceedings of the 30th International Conference on Advances in Geographic Information Systems}, pages 1--12, 2022.

\bibitem{safarinejadian2015consensus}
B.~Safarinejadian and M.~E. Estahbanati.
\newblock Consensus filter-based distributed variational bayesian algorithm for flow and speed density prediction with distributed traffic sensors.
\newblock {\em IEEE Systems Journal}, 11(4):2939--2948, 2015.

\bibitem{sengupta2021tqam}
R.~Sengupta, Y.~Karnati, A.~Rangarajan, and S.~Ranka.
\newblock Tqam: Temporal attention for cycle-wise queue length estimation using high-resolution loop detector data.
\newblock In {\em 2021 IEEE International Intelligent Transportation Systems Conference (ITSC)}, pages 3313--3320. IEEE, 2021.

\bibitem{tian2019hybrid}
X.~Tian, D.~Yu, X.~Xing, S.~Wang, and Z.~Wang.
\newblock Hybrid short-term traffic flow prediction model of intersections based on improved complete ensemble empirical mode decomposition with adaptive noise.
\newblock {\em Advances in Mechanical Engineering}, 11(4):1687814019841819, 2019.

\bibitem{velivckovic2017graph}
P.~Veli{\v{c}}kovi{\'c}, G.~Cucurull, A.~Casanova, A.~Romero, P.~Lio, and Y.~Bengio.
\newblock Graph attention networks.
\newblock {\em arXiv preprint arXiv:1710.10903}, 2017.

\bibitem{vigos2008real}
G.~Vigos, M.~Papageorgiou, and Y.~Wang.
\newblock Real-time estimation of vehicle-count within signalized links.
\newblock {\em Transportation Research Part C: Emerging Technologies}, 16(1):18--35, 2008.

\bibitem{wolf2014applying}
J.~L. Wolf, W.~H. Bachman, M.~S. Oliveira, J.~A. Auld, A.~Mohammadian, P.~S. Vovsha, and J.~Zmud.
\newblock {\em Applying GPS data to understand travel behavior}, volume~1.
\newblock Transportation Research Board, 2014.

\bibitem{wright2019neural}
M.~A. Wright, S.~F. Ehlers, and R.~Horowitz.
\newblock Neural-attention-based deep learning architectures for modeling traffic dynamics on lane graphs.
\newblock In {\em 2019 IEEE Intelligent Transportation Systems Conference (ITSC)}, pages 3898--3905. IEEE, 2019.

\bibitem{xia2016distributed}
D.~Xia, B.~Wang, H.~Li, Y.~Li, and Z.~Zhang.
\newblock A distributed spatial--temporal weighted model on mapreduce for short-term traffic flow forecasting.
\newblock {\em Neurocomputing}, 179:246--263, 2016.

\bibitem{yao2019cycle}
J.~Yao and K.~Tang.
\newblock Cycle-based queue length estimation considering spillover conditions based on low-resolution point detector data.
\newblock {\em Transportation research part C: emerging technologies}, 109:1--18, 2019.

\bibitem{yousefzadeh2024graph}
N.~Yousefzadeh, R.~Sengupta, Y.~Karnati, A.~Rangarajan, and S.~Ranka.
\newblock Graph attention network for lane-wise and topology-invariant intersection traffic simulation.
\newblock {\em arXiv preprint arXiv:2404.07446}, 2024.

\bibitem{yousefzadeh2023comprehensive}
N.~Yousefzadeh, M.~T~Thai, and S.~Ranka.
\newblock A comprehensive survey on multilayered graph embedding.
\newblock 2023.

\bibitem{zhao2019parallel}
L.~Zhao, Y.~Zhou, H.~Lu, and H.~Fujita.
\newblock Parallel computing method of deep belief networks and its application to traffic flow prediction.
\newblock {\em Knowledge-Based Systems}, 163:972--987, 2019.

\bibitem{zhao2019traffic}
S.~Zhao, Q.~Zhao, Y.~Bai, and S.~Li.
\newblock A traffic flow prediction method based on road crossing vector coding and a bidirectional recursive neural network.
\newblock {\em Electronics}, 8(9):1006, 2019.

\bibitem{zhao2019deep}
W.~Zhao, Y.~Gao, T.~Ji, X.~Wan, F.~Ye, and G.~Bai.
\newblock Deep temporal convolutional networks for short-term traffic flow forecasting.
\newblock {\em Ieee Access}, 7:114496--114507, 2019.

\end{thebibliography}

\end{document}

\section*{Supplementary Information}
\title{MTDT: A Multi-Task Deep Learning Digital Twin to Urban Intersections\\
% \thanks{The work was supported in part by NSF CNS 1922782 and by the Florida Department of transportation (FDOT). The opinions, findings and conclusions expressed in this publication are those of the authors and not necessarily those of NSF or FDOT.}
}

\makeatletter

\clearpage
\section{Supplementary Tables}
\vskip -2\baselineskip plus -1fil
\begin{table*}[htbp]
\centering
\begin{tabular}{llrrrr}
\toprule
 &  & EXT & INF & QLEN & TT \\
 &  & MAE & MAE & MAE & MAE\_percent \\
LEVEL & ISC &  &  &  &  \\
\midrule
\multirow[t]{8}{*}{L1} & J2 & 0.104 & 0.129 & 4.925 & 3.302 \\
 & J3 & 0.104 & 0.127 & 4.318 & 3.241 \\
 & J4 & 0.115 & 0.135 & 7.221 & 3.678 \\
 & J5 & 0.108 & 0.129 & 1.894 & 2.303 \\
 & J6 & 0.102 & 0.116 & 3.007 & 3.083 \\
 & J7 & 0.108 & 0.121 & 3.126 & 3.396 \\
 & J8 & 0.101 & 0.116 & 2.952 & 3.344 \\
 & J9 & 0.085 & 0.089 & 3.423 & 1.199 \\
\cline{1-6}
\multirow[t]{8}{*}{L2} & J2 & 0.103 & 0.130 & 6.004 & 3.076 \\
 & J3 & 0.103 & 0.127 & 4.588 & 2.956 \\
 & J4 & 0.112 & 0.138 & 7.466 & 3.437 \\
 & J5 & 0.107 & 0.133 & 4.505 & 1.936 \\
 & J6 & 0.102 & 0.121 & 6.019 & 2.979 \\
 & J7 & 0.107 & 0.127 & 6.998 & 2.506 \\
 & J8 & 0.100 & 0.118 & 5.316 & 3.092 \\
 & J9 & 0.083 & 0.091 & 4.533 & 1.008 \\
\cline{1-6}
\multirow[t]{8}{*}{M1} & J2 & 0.102 & 0.132 & 7.870 & 2.274 \\
 & J3 & 0.102 & 0.130 & 7.915 & 2.348 \\
 & J4 & 0.112 & 0.142 & 10.557 & 2.895 \\
 & J5 & 0.107 & 0.137 & 7.496 & 1.782 \\
 & J6 & 0.101 & 0.128 & 9.089 & 2.745 \\
 & J7 & 0.107 & 0.132 & 11.867 & 2.009 \\
 & J8 & 0.100 & 0.122 & 7.355 & 2.252 \\
 & J9 & 0.087 & 0.095 & 5.878 & 0.854 \\
\cline{1-6}
\multirow[t]{8}{*}{M2} & J2 & 0.101 & 0.133 & 10.362 & 1.642 \\
 & J3 & 0.101 & 0.130 & 12.198 & 2.291 \\
 & J4 & 0.113 & 0.145 & 14.925 & 2.446 \\
 & J5 & 0.108 & 0.139 & 11.493 & 1.852 \\
 & J6 & 0.103 & 0.133 & 11.062 & 2.758 \\
 & J7 & 0.106 & 0.133 & 16.684 & 1.892 \\
 & J8 & 0.099 & 0.126 & 9.770 & 1.597 \\
 & J9 & 0.088 & 0.097 & 6.960 & 0.800 \\
\cline{1-6}
\multirow[t]{8}{*}{H1} & J2 & 0.099 & 0.134 & 15.330 & 1.322 \\
 & J3 & 0.100 & 0.129 & 17.316 & 2.262 \\
 & J4 & 0.111 & 0.142 & 21.017 & 2.782 \\
 & J5 & 0.107 & 0.142 & 18.355 & 2.091 \\
 & J6 & 0.104 & 0.136 & 14.227 & 2.691 \\
 & J7 & 0.104 & 0.132 & 21.946 & 1.827 \\
 & J8 & 0.098 & 0.125 & 11.380 & 1.697 \\
 & J9 & 0.087 & 0.099 & 9.262 & 0.778 \\
\cline{1-6}
\multirow[t]{7}{*}{H2} & J2 & 0.093 & 0.126 & 30.028 & 1.098 \\
 & J3 & 0.101 & 0.127 & 19.146 & 2.631 \\
 & J4 & 0.111 & 0.140 & 15.539 & 2.774 \\
 & J6 & 0.105 & 0.131 & 10.209 & 2.993 \\
 & J7 & 0.103 & 0.131 & 31.597 & 1.773 \\
 & J8 & 0.100 & 0.123 & 9.484 & 2.138 \\
 & J9 & 0.085 & 0.094 & 12.895 & 0.802 \\
\cline{1-6}
\bottomrule
\end{tabular}

\caption{In this table, we see that the results (split based on Maximum Queue Length buckets) for various intersections when calculated separately, show consistent performance, indicating that our model is able to generalize across heterogeneous intersections. }

\end{table*}

\vskip -2\baselineskip plus -1fil
\begin{table*}[htbp]
\centering
\begin{tabular}{llrrrr}
\toprule
 &  & EXT & INF & QLEN & TT \\
 &  & MAE & MAE & MAE & MAE\_percent \\
LEVEL & ISC &  &  &  &  \\
\midrule
\multirow[t]{7}{*}{LOW} & J2 & 0.094 & 0.127 & 25.816 & 1.148 \\
 & J4 & 0.111 & 0.142 & 10.154 & 2.978 \\
 & J5 & 0.103 & 0.136 & 12.465 & 1.855 \\
 & J6 & 0.100 & 0.127 & 9.461 & 2.787 \\
 & J7 & 0.105 & 0.131 & 14.611 & 1.959 \\
 & J8 & 0.102 & 0.133 & 14.396 & 1.279 \\
 & J9 & 0.086 & 0.096 & 7.323 & 0.769 \\
\cline{1-6}
\multirow[t]{8}{*}{MEDIUM} & J2 & 0.101 & 0.133 & 13.940 & 1.702 \\
 & J3 & 0.105 & 0.140 & 9.149 & 1.674 \\
 & J4 & 0.112 & 0.140 & 10.850 & 3.015 \\
 & J5 & 0.109 & 0.138 & 8.016 & 1.833 \\
 & J6 & 0.103 & 0.125 & 7.411 & 2.847 \\
 & J7 & 0.108 & 0.130 & 10.956 & 2.317 \\
 & J8 & 0.102 & 0.127 & 7.481 & 2.315 \\
 & J9 & 0.087 & 0.094 & 5.790 & 0.846 \\
\cline{1-6}
\multirow[t]{8}{*}{HIGH} & J2 & 0.104 & 0.132 & 8.351 & 3.116 \\
 & J3 & 0.103 & 0.130 & 6.759 & 2.595 \\
 & J4 & 0.112 & 0.136 & 11.642 & 3.258 \\
 & J5 & 0.110 & 0.135 & 4.543 & 2.034 \\
 & J6 & 0.103 & 0.121 & 5.343 & 2.997 \\
 & J7 & 0.109 & 0.126 & 9.268 & 3.007 \\
 & J8 & 0.099 & 0.119 & 5.712 & 2.779 \\
 & J9 & 0.086 & 0.092 & 5.849 & 1.197 \\
\cline{1-6}
\bottomrule
\end{tabular}
\caption{In this table, we see that the results (split based on Corridor-Through Green Time Percentage buckets) for various intersections when calculated separately, show consistent performance, indicating that our model is able to generalize across heterogeneous intersections.}

\end{table*}